\documentclass{Interspeech}

\interspeechcameraready

\title{LitMAS: A Lightweight and Generalized Multi-Modal Anti-Spoofing Framework for Biometric Security}

\author[affiliation={1}]{Nidheesh}{Gorthi}
\author[affiliation={2}]{Kartik}{Thakral}
\author[affiliation={2}]{Rishabh}{Ranjan}
\author[affiliation={2}]{Richa}{Singh}
\author[affiliation={2}]{Mayank}{Vatsa}

\affiliation{}{Indian Institute of Information Technology Kottayam}{India}
\affiliation{}{Indian Institute of Technology Jodhpur}{India}
\email{nidheesh21bcs143@iiitkottayam.ac.in, \{thakral.1, ranjan.4,  richa, mvatsa\}@iitj.ac.in}


\usepackage{comment}
\usepackage{multirow}
\usepackage{pifont}
\usepackage{subcaption}
\usepackage{footnote}
\makesavenoteenv{tabular} 
\usepackage{tablefootnote}
\usepackage{hyperref}

\begin{document}

\maketitle

\begin{abstract}
    

    \noindent Biometric authentication systems are increasingly being deployed in critical applications, but they remain susceptible to spoofing. Since most of the research efforts focus on modality-specific anti-spoofing techniques, building a unified, resource-efficient solution across multiple biometric modalities remains a challenge. To address this, we propose LitMAS, a \textbf{Li}gh\textbf{t} weight and generalizable \textbf{M}ulti-modal \textbf{A}nti-\textbf{S}poofing framework designed to detect spoofing attacks in speech, face, iris, and fingerprint-based biometric systems. At the core of LitMAS is a Modality-Aligned Concentration Loss, which enhances inter-class separability while preserving cross-modal consistency and enabling robust spoof detection across diverse biometric traits. With just 6M parameters, LitMAS surpasses state-of-the-art methods by 1.36\% in average EER across seven datasets, demonstrating high efficiency, strong generalizability, and suitability for edge deployment. Code and trained models are available at \url{https://github.com/IAB-IITJ/LitMAS}.
\end{abstract}

\section{Introduction}
Biometric authentication has become a cornerstone of secure identity verification, widely used in banking, mobile authentication, access control, and emerging technologies such as AR/VR systems and smart assistants. These systems leverage speech, face, iris, and fingerprint recognition, which can be used individually or in combination to enhance security. For instance, speech-based authentication is common in voice assistants and telephone banking, while face recognition is used in smartphone unlocking and airport security. Iris recognition is applied in high-security environments like border control, and fingerprint authentication is a standard method for financial transactions and personal device security. In multi-modal authentication systems, such as those used in financial institutions, combining multiple biometric traits further strengthens the security.

\par
\begin{figure}[t]
    \centering
    \includegraphics[width=1.0\columnwidth]{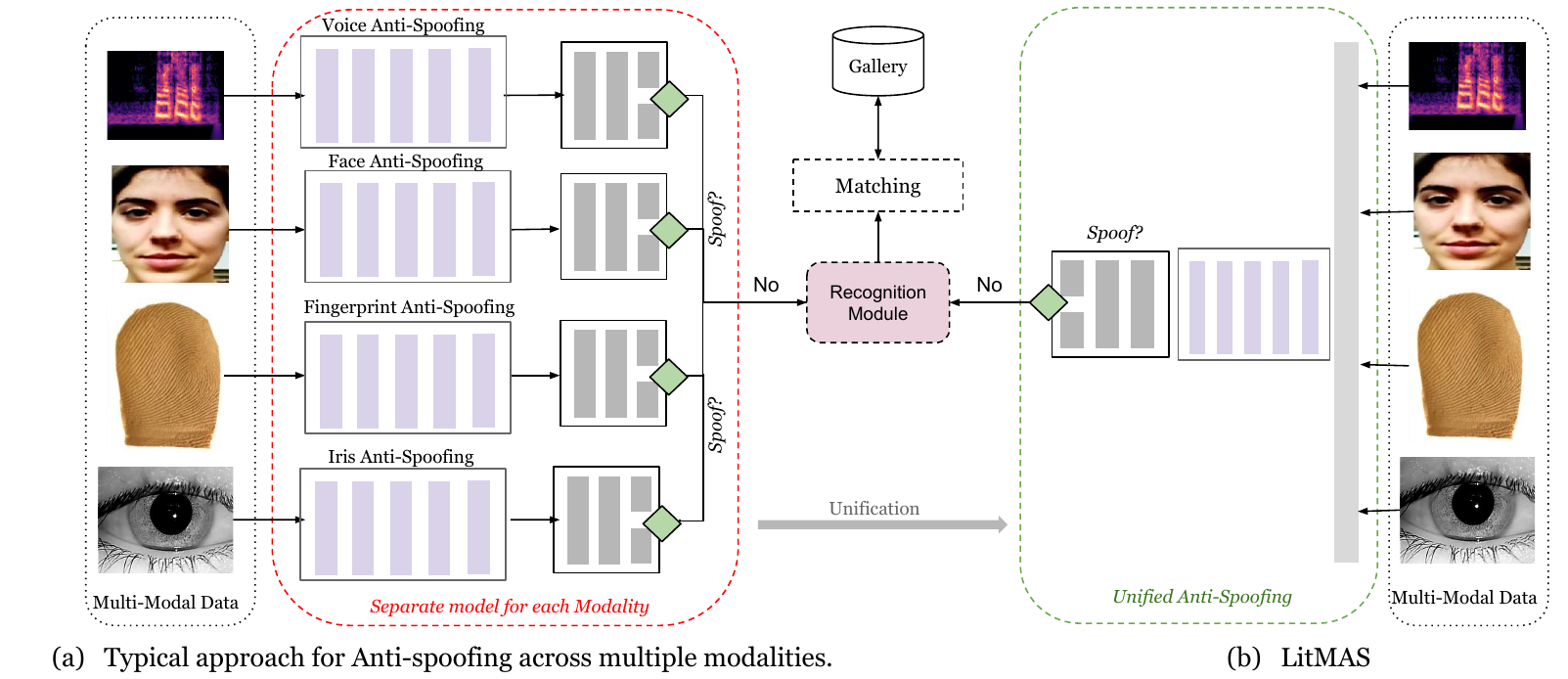}
    \caption{Comparison of (a) existing multi-modal anti-spoofing approach and (b) a unified approach (proposed).}
    \label{fig:VisAbstract}
\end{figure}

 Despite their widespread adoption, biometric systems remain highly susceptible to spoofing attacks, where adversaries attempt to bypass authentication using fake biometric traits, such as replayed voice recordings, printed face images, synthetic irises, or fabricated fingerprints \cite{thakral2023phygitalnet}. While extensive research has been conducted on modality-specific anti-spoofing solutions \cite{faceantisp, witkowski2017audio, fang2020robust, fingpad}, real-world biometric authentication increasingly relies on multi-modal systems where existing methods fail, creating a demand for generalized anti-spoofing methods (as illustrated in Figure \ref{fig:VisAbstract}) that can generalize across different biometric traits. Furthermore, deploying anti-spoofing models on resource-constrained edge devices such as smartphones, IoT systems, and AR glasses remains a challenge, as most existing solutions are computationally expensive and lack efficiency.

\begin{figure*}[t]
    \centering
    \includegraphics[width=1\textwidth]{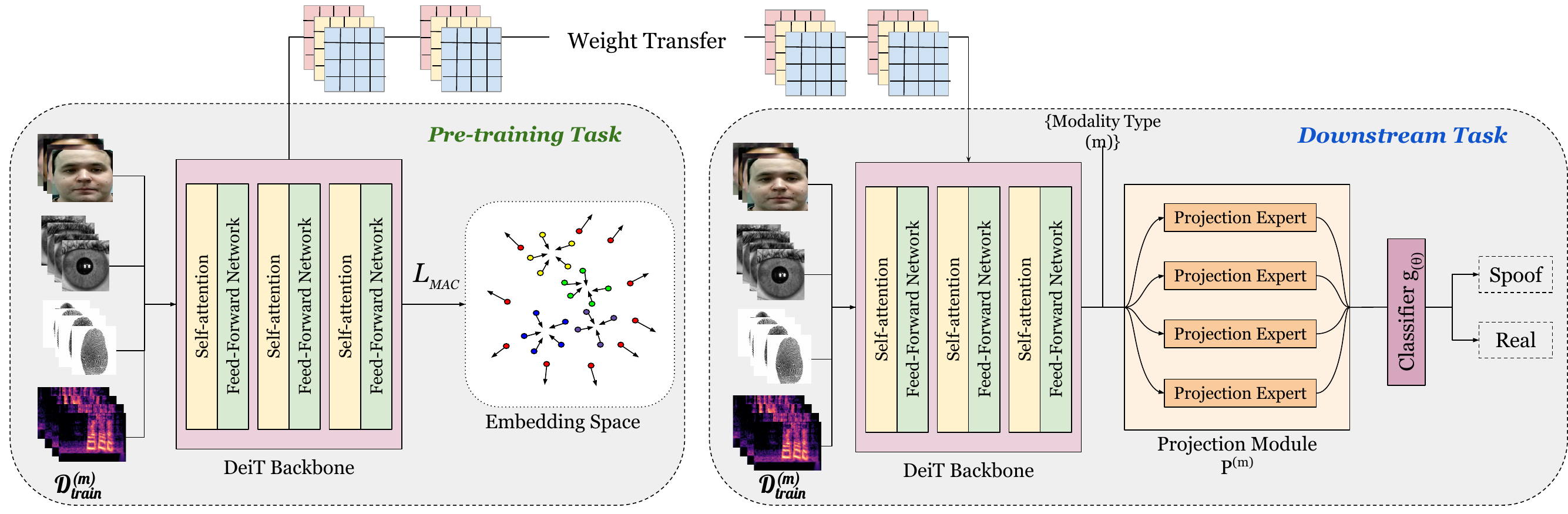}
    \caption{Illustration of the proposed LitMAS architecture (best viewed in color). The left block demonstrates the first step of the pre-training task with the MAC loss, and the right block demonstrates the second step of MoPE architecture used for the downstream~task.}
    \label{fig:ModelArchitecture}
\end{figure*}

\par Over the years, substantial progress has been made in anti-spoofing research across individual biometric modalities. In voice anti-spoofing, techniques such as linear frequency cepstral coefficients (LFCC) and constant-Q cepstral coefficients (CQCC) \cite{todisco16_odyssey} have been widely used. At the same time, recent approaches focus on processing raw audio \cite{ranjan2024context, ranjan2023sv} and data augmentation \cite{sanchez24_iberspeech} to further improve detection efficiency. Face anti-spoofing has evolved to counter various attacks, including printed photos, video replays, and 3D masks, with recent advancements leveraging hyperbolic embeddings \cite{10658085}, Vision Transformers \cite{AAViT}, and lightweight CNN models \cite{Luevano_FAS_CVPR2024} optimized for real-time detection. Recent digital attacks, particularly face-swapping techniques, have been extensively studied in the literature \cite{thakralillusion, narayan2023df, narayan2022deephy}, along with the development of corresponding defense mechanisms \cite{thakral2024deephynet, chhabra2023low, narayan2022desi}. Similarly, iris anti-spoofing techniques have progressed from handcrafted features such as LBP and HOG \cite{9780578} to deep learning-based classification methods with different backbones \cite{mitcheff2024privacysafeirispresentationattack}. Recent works in fingerprint anti-spoofing have transitioned from traditional Gabor and LBP descriptors to deep learning solutions, including Transformer-based classifiers \cite{10449087} and MobileNet-SVM architectures~\cite{RAI2025104069}.

\par Despite advancements in modality-specific anti-spoofing, existing methods struggle with cross-modal generalization and are often computationally expensive, limiting their deployment on real-world edge devices. To address this, we propose LitMAS, a generalized and lightweight anti-spoofing framework that works across multiple biometric modalities. Our key contributions are as follows:

\begin{itemize}
    \item A Unified Anti-Spoofing Framework termed LitMAS, which is the first multi-modal anti-spoofing model. It is capable of detecting attacks on speech, face, iris, and fingerprint with a single, efficient architecture, while processing a single modality at a time.
    \item We propose Modality-Aligned Concentration (MAC) Loss to enhance separation between live and spoof samples while preserving modality-specific information. Additionally, our Mixture of Projection Experts (MoPE) architecture balances modality-invariant representations with modality-aware feature retention, ensuring robust generalization.
    \item With only $\sim$6M parameters, LitMAS achieves state-of-the-art anti-spoofing performance while being resource-efficient for real-time edge deployments. We validate its effectiveness across seven diverse datasets, demonstrating superior accuracy and generalization over existing methods.

\end{itemize}

\begin{table*}[t]
  \caption{AUC (\(\uparrow\)) and EER (\%) (\(\downarrow\)) values reported across the four biometric modalities for performance comparison of different baseline algorithms and the proposed LitMAS algorithm. Each baseline is re-trained on all the modalities.}
  \centering
  \begin{tabular}{l c c c c c c c c c c}
    \toprule
    \multirow{2}{*}{\textbf{Algorithm}} & \multicolumn{2}{c}{\textbf{Speech}} & \multicolumn{2}{c}{\textbf{Iris}} & \multicolumn{2}{c}{\textbf{Face}} & \multicolumn{2}{c}{\textbf{Fingerprint}} & \multicolumn{2}{c}{\textbf{Average}} \\
    \cmidrule(r){2-3} \cmidrule(r){4-5} \cmidrule(r){6-7} \cmidrule(r){8-9} \cmidrule(r){10-11}
    & \textbf{AUC}  & \textbf{EER}  & \textbf{AUC}  & \textbf{EER}  & \textbf{AUC}  & \textbf{EER}  & \textbf{AUC}  & \textbf{EER}  & \textbf{AUC} & \textbf{EER} \\
    \midrule
    LCNN \cite{wang21fa_interspeech}       & 0.9846  & 6.43  & 0.9921  & 4.47  & 0.9295  & 14.14  & 0.9139  & 16.51  & 0.9550  & 10.39  \\
    SpecRNet \cite{10063734}        & 0.9780  & 7.46  & 0.9872  & 6.30  & 0.9412  & 12.73  & 0.9629  & 9.78   & 0.9673  & 9.07   \\
    ViTUnified   \cite{10449087}       & 0.9680  & 9.25  & 0.9626  & 11.97 & 0.9498  & 12.09  & 0.8528  & 23.03  & 0.9333  & 14.09 \\
    MoSFPAD    \cite{RAI2025104069}         & 0.9544  & 11.91 & \textbf{0.9967}  & \underline{3.25}  & \underline{0.9863}  & \underline{6.31}   & 0.9809  & 6.84   & 0.9796  & 7.08   \\
    MobileNetV3-Spoof \cite{Luevano_FAS_CVPR2024}   & 0.9427  & 13.20 & 0.9945  & 3.83  & 0.9748  & 8.08   & \underline{0.9830}  & \textbf{6.65}   & 0.9738  & 7.94   \\
    AAViT \cite{AAViT}       & \underline{0.9898}  & \underline{4.64}  & 0.9936  & 3.83  & 0.9768  & 8.07   & 0.9752  & 8.37   & \underline{0.9839}  & \underline{6.23}   \\
    CascadeOfNetworks \cite{Cascade}   & 0.9461  & 13.00 & 0.9863  & 6.50  & 0.9736  & 9.02   & 0.9650  & 10.12  & 0.9678  & 9.66   \\
    \midrule  
    LitMAS (Proposed)    & \textbf{0.9916}  & \textbf{4.54}  & \underline{0.9962}  & \textbf{3.13}  & \textbf{0.9894}  & \textbf{5.06}   & \textbf{0.9837}  & \underline{6.70}   & \textbf{0.9902}  & \textbf{4.86}  \\
    \bottomrule
  \end{tabular}
  \label{tab:big_table}
\end{table*}

\section{Proposed LitMAS Framework for Multi-Modal Anti-Spoofing}
\label{sec:method}

\noindent The proposed framework is illustrated in Figure \ref{fig:ModelArchitecture}. Let $\mathcal{M} = \{m_1, m_2, \dots, m_{|\mathcal{M}|}\}$ be the set of biometric modalities we aim to protect against spoofing (e.g., speech, face, iris, fingerprint). Let the combined training set be defined as
\begin{equation}
    \mathcal{D}_{\text{train}} = \displaystyle\bigcup_{m \in \mathcal{M}} \mathcal{D}_{\text{train}}^{(m)}
\end{equation}
where $\mathcal{D}_{\text{train}}^{(m)}$ is the training subset for modality $m$. Each sample $x_i$ in $\mathcal{D}_{\text{train}}^{(m)}$ is associated with a label $y_i \in \{0,1\}$, where $y_i = 0$ indicates a \emph{bonafide} (real) sample and $y_i = 1$ indicates a \emph{spoofed} sample. Our goal is to learn a single unified anti-spoofing model, \text{LitMAS}, which processes input samples $x_i$ from any modality $m$ and reliably classifies them into bonafide or spoof. To achieve strong cross-modal discriminability while remaining lightweight, LitMAS proceeds in two main steps: a pre-training step with a novel \emph{Modality-Aligned Concentration (MAC) Loss} and a fine-tuning step that employs a \emph{Mixture of Projection Experts (MoPE)} architecture.

\noindent\textbf{Step 1: Pre-training with Modality-Aligned Concentration (MAC) Loss:} A key objective in multi-modal anti-spoofing is to obtain an embedding space in which: (i) Bonafide samples of the same modality are pulled closer together. (ii) Bonafide samples of different modalities still maintain class consistency. (iii) Spoof samples (across \textit{all} modalities) are pushed farther from \textit{all} bonafide embeddings.

Let $\mathbf{z}_i$ be the embedding of sample $x_i$ extracted by an initial backbone network. For each modality $m$, let $\mathbf{C}^m$ be the \emph{center} (prototype) of bonafide embeddings for that modality. After each epoch, $\mathbf{C}^m$ is updated to be the mean of the bonafide embeddings currently assigned to modality $m$. Formally, let $N_m$ be the set of indices corresponding to (i) bonafide samples from modality $m$ and (ii) spoof samples from any modality in the current training batch. We introduce the MAC loss in two stages. First, for modality $m$, we compute the concentration loss $L_c$ as:
\begin{equation}
\label{eq:lc}
L_{c}^m = -\frac{1}{|N_m|} \sum_{i \in N_m} \hat{y}_i \log\Biggl(\frac{\exp\bigl(s^m_i\bigr)}{\sum_{j \in N_m} \exp\bigl(s^m_j\bigr)}\Biggr).
\end{equation}
where $s^m_i$ is the cosine similarity between $\mathbf{z}_i$ and the modality center $\mathbf{C}^m$, and $\hat{y}_i$ is the softmax-transformed label for sample i. This loss is designed to \emph{pull} bonafide embeddings of modality $m$ toward $\mathbf{C}^m$, while pushing away spoof samples of any modality. Next, we average this concentration loss over all modalities in the batch to define MAC loss as:
\begin{equation}
\label{eq:mac}
L_{\text{MAC}} = \frac{1}{|\mathcal{M}|}\sum_{m \in \mathcal{M}} L_{c}^m.
\end{equation}

This ensures that every modality jointly contributes to separating real versus spoof embeddings while also preserving intra-class consistency for each modality’s bonafide samples. The initial pre-training step establishes a robust initialization for the embedding space.

\vspace{6pt}
\noindent\textbf{Step 2: Mixture of Projection Experts (MoPE):} Although the pre-trained backbone now produces representations that partially separate bonafide and spoof samples, each modality might still exhibit unique cues (e.g., replay artifacts in speech vs. texture anomalies in fingerprint images). To incorporate these modality-specific nuances, we introduce a \emph{Mixture of Projection Experts (MoPE)} layer.

Concretely, we define a set of projection heads $\bigl\{\mathbf{P}^{(m)}: \mathbb{R}^d \to \mathbb{R}^{k}\bigr\}_{m \in \mathcal{M}}$, with $d$ as the dimension of the backbone’s embedding space, and $k$ as a (possibly) larger feature dimension that better captures the fine-grained cues needed to discriminate spoof attacks for modality $m$. During training, a sample $x_i$ from modality $m$ is projected through the corresponding head $\mathbf{P}^{(m)}$. In this way, each projection expert captures the critical, modality-specific features necessary for classification.

After aggregating these transformed features, we utilize a lightweight classifier $g_\theta(\cdot)$ to yield the final spoof vs. bonafide score. The complete model is then trained end-to-end using a standard cross-entropy objective on top of the pre-trained weights, thereby allowing both the backbone and the projection experts to refine the embeddings jointly.

\begin{table}[t]
    \centering
    \caption{Performance comparison of baseline algorithms using min t-DCF for speech modality, and BPCER @ APCER 1\% for iris, face, and fingerprint modalities.}
    \resizebox{\columnwidth}{!}{%
    \begin{tabular}{lcccc}
        \toprule
        \multirow{2}{*}{\textbf{Algorithm}} & \textbf{Speech} & \textbf{Iris} & \textbf{Face} & \textbf{Fingerprint} \\
        \cmidrule(r){2-2} \cmidrule(l){3-5}
        & \textbf{min t-DCF} & \multicolumn{3}{c}{\textbf{BPCER @ APCER 1\%}} \\
        \midrule
        LCNN \cite{wang21fa_interspeech}& 0.16033 & 12.09 & 68.18 & 70.08 \\
        SpecRNet \cite{10063734}& 0.18505 & 18.58 & 100.00 & 51.31 \\
        ViTUnified \cite{10449087}& 0.23985 & 25.2 & 44.57 & 86.43 \\
        MosFPAD \cite{RAI2025104069}& 0.26278 & \textbf{6.04} & \underline{17.55} & 32.05 \\
        MobileNetV3 \cite{Luevano_FAS_CVPR2024}& 0.31878 & 9.78 & 27.25 & \underline{28.77} \\
        AAViT \cite{AAViT}& \textbf{0.11234} & 10.09 & 31.15 & 31.44 \\
        Cascade \cite{Cascade}& 0.29743 & 16.40 & 32.40 & 47.84 \\
        \midrule
        LitMAS  & \underline{0.11585} & \underline{6.36} & \textbf{12.58} & \textbf{25.5} \\
        \bottomrule
    \end{tabular}%
    }
    \label{tab:biometric_performance}
\end{table}

\vspace{6pt}
\noindent\textbf{LitMAS:} Our framework follows a two-step learning strategy. First, we pre-train with the MAC loss to learn well-separated embeddings that cluster bonafide samples by modality and push spoof samples away (see left block of Figure~\ref{fig:ModelArchitecture}). Next, we introduce projection experts $\mathbf{P}^{(m)}$ and a final classifier, refining for each modality in an end-to-end fashion (right block). Each expert captures fine-grained, modality-specific cues while leveraging the robust, globally discriminative embeddings from the first step. This synergy delivers strong cross-modal generalization at minimal computational cost, making \textit{LitMAS} ideal for resource-constrained, edge devices.

\section{Experimental Setup}
\noindent\textbf{Datasets and Pre-processing:} To evaluate the generalizability of our approach, we conduct experiments on diverse datasets spanning multiple biometric modalities, covering various types of spoofing attack. Statistics for all the utilized datasets are presented in Table \ref{tab:class_distribution}.

\begin{table}[!t]
    \centering
   \caption{Class distribution of real and spoof samples across datasets. A total of 64,382 samples were used for training and 176,875 samples were used for testing.}
   \resizebox{\columnwidth}{!}{
    \begin{tabular}{lcccc}
        \toprule
        \multirow{2}{*}{\textbf{Dataset}} & \multicolumn{2}{c}{\textbf{Train}} & \multicolumn{2}{c}{\textbf{Test}} \\
        \cmidrule(lr){2-3} \cmidrule(lr){4-5}
        & \textbf{Real} & \textbf{Spoof} & \textbf{Real}  & \textbf{Spoof}  \\
        \midrule
        ASVSpoof-2019 \cite{wu15e_interspeech}             & 5,400  & 14,040 & 18,090 & 116,640 \\
        Iris CSD \cite{iriscsd}            & 7,813  & 6,750  & 2,605  & 2,250  \\
        LivDet Fingerprint \cite{mura2018livdet}         & 5,998  & 7,198  & 5,092  & 6,087  \\
        MSU-MFSD \cite{msumfsd}          & 900   & 2,683  & 1,200  & 3,592  \\
        Replay Attack \cite{replayattack}    & 1,800  & 8,994  & 2,400  & 12,000 \\
        Silicone Mask \cite{silicone}       & 237   & 666   & 501   & 664   \\
        MLFP \cite{mlfp}            & 476   & 1,427  & 720   & 5,034  \\
        \midrule
        Total     & 22,624 & 41,758 & 30,608 & 146,267 \\
        \bottomrule
    \end{tabular}}
 
    \label{tab:class_distribution}
\end{table}
\begin{table}[t]
  \caption{APCER and BPCER values (at EER threshold) reported for the datasets used.}
  \centering
  \begin{tabular}{l c c}
    \toprule
    \textbf{Dataset}    & \textbf{APCER} (\%) & \textbf{BPCER} (\%) \\
    \midrule
    ASVSpoof 2019 \cite{wu15e_interspeech}   & 4.538   & 4.541   \\
    Iris CSD \cite{iriscsd}        & 3.129   & 3.147   \\
    LivDet-2017 Fingerprint \cite{mura2018livdet}    & 6.696   & 6.702   \\
    MSU-MFSD \cite{msumfsd}   & 5.750   & 5.818   \\
    Replay Attack \cite{replayattack}    & 0.416   & 0.449   \\
    Silicone Mask \cite{silicone}  & 16.367  & 16.566  \\
    MLFP \cite{mlfp}      & 13.055  & 12.991  \\
    \bottomrule
  \end{tabular}
\label{tab:apcer_bpcer}
\end{table}

\begin{figure*}[t]
    \centering
    \includegraphics[width=\textwidth]{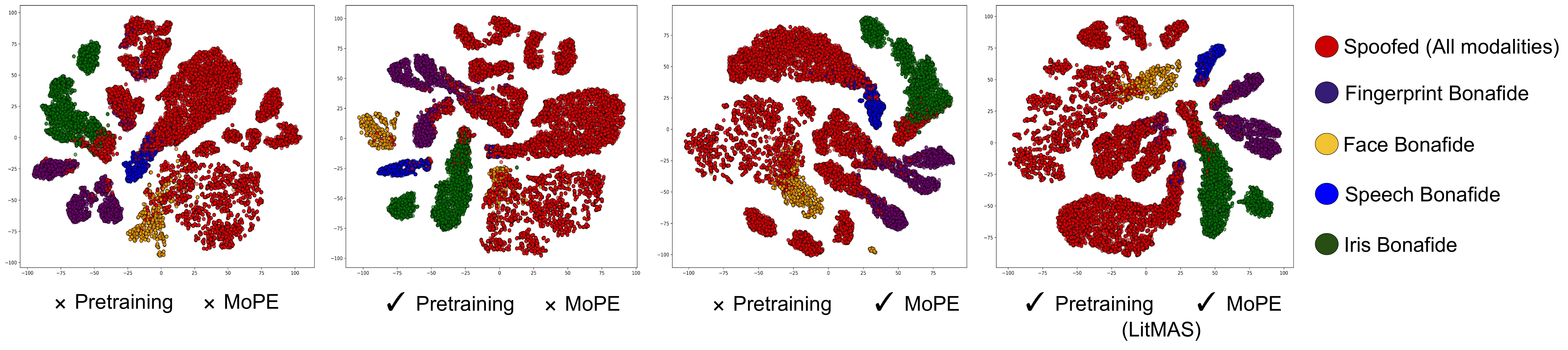}
    \caption{t-SNE visualizations depicting the separation between spoofed and bonafide samples across different ablation~configurations.}
    \label{fig:tsne}
\end{figure*}


\vspace{6pt}
\noindent\textbf{Baseline Methods:} We evaluate LitMAS against state-of-the-art methods across individual biometric modalities. To ensure a fair comparison, all baselines are trained on the combined multi-modal dataset and evaluated per modality, following our unified training protocol. We include strong baselines that are representative state-of-the-art methods for each modality: LCNN \cite{wang21fa_interspeech}, SpecRNet \cite{10063734}, AAViT \cite{AAViT}, MobileNetV3-spoof \cite{Luevano_FAS_CVPR2024}, CascadeOfNetworks \cite{Cascade}, ViTUnified \cite{10449087} and MoSFPAD \cite{RAI2025104069}. 

\vspace{6pt}
\noindent\textbf{Evaluation Metrics:} We evaluate performance using AUC and Equal Error Rate (EER). EER is the threshold where APCER (spoof misclassified as live) and BPCER (live misclassified as spoof) are equal, offering a balanced assessment, especially for imbalanced datasets. For speech, we also report the minimum tandem detection cost function (min t-DCF) to evaluate the trade-off between missed detections and false alarms. 

\vspace{6pt}
\noindent\textbf{Implementation details:} We use DeiT-Tiny\footnote{https://github.com/facebookresearch/deit} \cite{touvron2021training} as the backbone for its efficiency and lightweight design. The model is trained on the combined dataset across all modalities and evaluated per modality. We employ the AdamW optimizer (learning rate: 1e-4, weight decay: 1e-5) with a batch size of 64, ensuring balanced bonafide samples in each batch. In Step 1, the [CLS] token’s feature size is 192, which is projected to 512 in Step 2 using modality-specific projection experts. Training is carried out for 40 epochs in both steps. We modify the input layer of SpecRNet \cite{10063734} to accept 3-channel inputs, and DeiT-Tiny is used as the backbone for AAViT \cite{AAViT}. Experiments are conducted on an Nvidia A40 and three A30 GPUs (48GB and 24GB VRAM,~respectively).

\section{Results and Discussion}
\vspace{-2pt}

\noindent\textbf{Generalization on Multi-Modal Anti-Spoofing Detection:} Table \ref{tab:big_table} summarizes the detection performance of the proposed \textit{LitMAS} framework compared against state-of-the-art methods in terms of AUC and EER across four biometric modalities: speech, iris, face, and fingerprint. As shown, \textit{LitMAS} consistently outperforms all baselines, achieving superior average AUC and EER. In particular, it demonstrates the largest performance gains in the face and fingerprint modalities, where existing approaches typically suffer due to modality-specific artifacts and limited generalization. 

We observe that certain methods like MoSFPAD \cite{RAI2025104069} and MobileNetV3-Spoof \cite{Luevano_FAS_CVPR2024} excel on individual modalities but exhibit lower performance on others. For instance, MoSFPAD shows strong results on iris spoof detection but underperforms on speech-based attacks. Similarly, MobileNetV3-Spoof works well on fingerprint attacks but fails to generalize effectively to speech spoofs. By contrast, \textit{LitMAS} provides consistent improvement across \emph{all} modalities, highlighting its superior multi-modal learning strategy. 

In Table \ref{tab:biometric_performance}, we also report the min t-DCF for the speech modality, and BPCER@1\% APCER for the remaining three modalities. The BPCER@1\% APCER measures BPCER when APCER is fixed at 1\%. This metric is particularly useful because it evaluates the performance of the system under a strict false acceptance constraint. In addition to AUC and EER, we computed APCER and BPCER for a more fine-grained analysis. Table \ref{tab:apcer_bpcer} reports the values for each dataset at the EER threshold. Our framework not only sustains a low APCER (i.e., fewer spoofed samples are misclassified as real) but also keeps the BPCER low (fewer real samples misclassified as spoof). This is crucial for practical settings since a system with low APCER but very high BPCER would excessively flag legitimate attempts as spoofs, reducing usability.




\noindent\textbf{Ablation Study and t-SNE Visualization:}
Table~\ref{tab:ablation} compares four configurations that isolate the effects of MAC pre-training and MoPE. Without either component, the model reaches an AUC of 0.9843 and an EER of 6.17\%. Adding MAC or MoPE individually provides a modest boost (AUC $\approx$ 0.986, EER $\approx$ 5.7\%), while combining both yields the highest AUC (0.9902) and lowest EER (4.86\%), confirming their synergistic benefit for multi-modal spoof detection. Figure~\ref{fig:tsne} illustrates this progression via t-SNE embeddings: models lacking MAC or MoPE exhibit substantial overlap between bona fide and spoof points, whereas \textit{LitMAS} (rightmost panel) produces well-separated clusters across all modalities. This visual evidence aligns with the quantitative results, reinforcing the effectiveness of our proposed design.

\noindent\textbf{Discussion}
Overall, the results demonstrate that \textit{LitMAS} addresses the central challenges of multi-modal anti-spoofing by learning a structured embedding space and refining it with modality-specific projection experts, enabling strong cross-modal generalization even against widely varying spoof artifacts (e.g., synthetic speech vs. silicone masks). Despite being lightweight (with $\sim$6M parameters), the framework maintains state-of-the-art performance and remains practical for real-time, on-device deployment. Additionally, the ablation study and dataset-specific metrics confirm \textit{LitMAS}’s adaptability in handling diverse attacks. These findings highlight the potential of \textit{LitMAS} as a robust, resource-efficient anti-spoofing solution that bridges the gap between stringent security requirements and the computational constraints of modern biometric applications.

\begin{table}[t]
\caption{Ablation study results for different training combinations of pre-training with MAC loss and the MoPE architecture}
    \centering
    \begin{tabular}{c c c c}
        \hline
        \textbf{Pretraining} & \textbf{MoPE} & \textbf{AUC} & \textbf{EER} \\
        \hline
        \ding{55} & \ding{55} & 0.9843 & 6.17 \\
        \ding{51} & \ding{55} & 0.9862 & 5.64 \\
        \ding{55} & \ding{51} & 0.9864 & 5.75 \\
        \ding{51} & \ding{51} &\textbf{0.9902} & \textbf{4.86} \\
        \hline
    \end{tabular}
    \label{tab:ablation}
\vspace{-5pt}
\end{table}

\vspace{-2pt}
\section{Conclusion}
\vspace{-2pt}
We presented \textit{LitMAS}, a unified anti-spoofing framework that leverages a novel Modality-Aligned Concentration (MAC) loss and a Mixture of Projection Experts (MoPE) to jointly handle speech, face, iris, and fingerprint attacks. Through extensive experiments, we demonstrated that LitMAS not only disentangles modality-specific features, but also ensures robust live-spoof separation, leading to consistent gains over state-of-the-art methods. Crucially, the approach remains lightweight, making it well-suited for resource-constrained edge devices. Overall, LitMAS offers a scalable, high-performance solution to multi-modal anti-spoofing, closing the gap between stringent security requirements and practical deployment challenges.

\vspace{-2pt}
\section{Acknowledgement}
\vspace{-2pt}

This research is supported by a grant from IndiaAI and Meta via the Srijan: Centre of Excellence for Generative AI. Gorthi was supported by ACM IKDD Uplink Internship and Thakral by the PMRF Fellowship.

\bibliographystyle{IEEEtran}
\bibliography{mybib}

\end{document}